\documentclass[10pt,letterpaper,oneside,twocolumn]{IEEEtran}

\usepackage{graphicx}
\usepackage{amsmath}
\usepackage{amsfonts}
\usepackage{epsfig}
\usepackage{url}
\usepackage{cite}

\newcommand{\BibTeX}{{\rmfamily B\kern-.05em{\scshape i\kern-.025em
b}\kern-.08em T\kern-.1667em\lower.7ex\hbox{E}\kern-.125emX}}

\renewcommand{\baselinestretch}{1}

\begin{document}

\title{A New Approach to Automated Epileptic Diagnosis Using EEG and Probabilistic Neural Network}

\author{
Forrest Sheng Bao$^{1,2}$ , Donald Yu-Chun Lie$^{2}$, and Yuanlin Zhang$^{1}$ \\
$^1$ Dept. of Computer Science  $^2$ Dept. of Electrical and Computer Engineering\\
Texas Tech University, Lubbock, Texas, USA 

 \thanks{This research is supported by the Fall 2007 Research Enrichment Fund Grant of Texas Tech University.
\newline E-mails: forrest.bao @gmail.com, donald.lie @ttu.edu, yzhang @cs.ttu.edu
 }}
\maketitle

\begin{abstract}
Epilepsy is one of the most common neurological disorders that greatly impair patient' daily lives. Traditional epileptic diagnosis relies on tedious visual screening by neurologists from lengthy EEG recording that requires the presence of seizure (ictal) activities. Nowadays, there are many systems helping the neurologists to quickly find interesting segments of the lengthy signal by automatic seizure detection. However, we notice that it is very difficult, if not impossible, to obtain long-term EEG data with seizure activities for epilepsy patients in areas lack of medical resources and trained neurologists. Therefore, we propose to study automated epileptic diagnosis using interictal EEG data that is much easier to collect than ictal data. The authors are not aware of any report on automated EEG diagnostic system that can accurately distinguish patients' interictal EEG from the EEG of normal people. The research presented in this paper, therefore, aims to develop an automated diagnostic system that can use interictal EEG data to diagnose whether the person is epileptic. Such a system should also detect seizure activities for further investigation by doctors and potential patient monitoring. To develop such a system, we extract four classes of features from the EEG data and build a Probabilistic Neural Network (PNN) fed with these features. Leave-one-out cross-validation (LOO-CV) on a widely used epileptic-normal data set reflects an impressive 99.5\% accuracy of our system on distinguishing normal people's EEG from patient's interictal EEG. We also find our system can be used in patient monitoring (seizure detection) and seizure focus localization, with 96.7\% and 77.5\% accuracy respectively on the data set. 
\end{abstract}

\begin{IEEEkeywords}
Epilepsy, Electroencephalogram (EEG), Probabilistic Neural Network (PNN), seizure.
\end{IEEEkeywords}

\section{Introduction}

\IEEEPARstart{E}{pilepsy} is a chronic disorder characterized by recurrent seizures, which may vary from muscle jerks to severe convulsions~\cite{Gastaut1973}. Estimated 1\% of world population suffers from epilepsy~\cite{BME_Prediction}, while 85\% of them live in the developing countries~\cite{WHO_Atlas}. Electroencephalogram (EEG) is routinely used clinically to diagnose, monitor and localize epileptogenic zone~\cite{purpose_of_EEG}. Long-term EEG monitoring can provide 90\% positive diagnostic information \cite{90Positive} and thus become a golden standard in epilepsy diagnosis. 

Traditional methods rely on experts to visually inspect the entire lengthy EEG recordings of up to 1 week, which is tedious and time-consuming. Therefore, many automated system assisting the diagnosis of epilepsy have emerged~\cite{ One-Class_Novelty_Detection_for_Seizure_Analysis_from_Intracranial_EEG,1084730,243992,A_neural-network-based_detection_of_epilepsy,1028922,lvq,4167902}. They could detect abnormal EEG segments related to seizures so that doctors can quickly view events of interest without having to page through the entire recording~\cite{gorgan99}. But this approach requires the presence of seizure activities in the EEG data. This tough requirement often leads to very long, even up to 1 week, continuous EEG recording to capture seizure activities because of the difficulty to tell if and when a seizure will occur. The long-term EEG recording can greatly disturb patients' daily lives. Another clinical concern is that very unfortunetaly, 50-75\% epilepsy patients in the world reside in areas lack of medical resources and trained professionals \cite{WHO_Atlas}, which makes the long-term EEG recording virtually impractical to those people. Therefore, an automated EEG epilepsy diagnostic system would be very valuable if it does not require data from active seizure activities (i.e., ictal) to perform the diagnosis. However, to the authors' best knowledge, we are not aware of any report on automated epilepsy diagnostic system using only interictal EEG data. 

In this paper, we aim to develop an automated system that can diagnose epilepsy not only by using ictal EEG data but also interictal data.  The diagnosis function of this system will be valuable for patients in areas lack of medical resources, and particularly well-trained personnels. Its capability of seizure detection will be the base of effective monitoring in personal health care and help doctors to do further diagnosis if necessary. In addition to diagnosis and seizure detection, we would also like the system to provide basic information on focus localization which is also an important aspect in diagnosis. 

Our system is a Probabilistic Neural Network (PNN)~\cite{PNN} based classifier. Previous research suggests PNN is more suitable for medical applications, since it uses Bayesian strategies, a process familiar to medical decision makers~\cite{MedPNN}. We adopt PNN for its fast speed, high accuracy and real-time property in updating network structure, as we will explain in Sec.~\ref{PNN}.
It is very difficult to directly use raw EEG data as input to an Artificial Neural Network~\cite{NTUpaper}. Therefore, a key in designing the PNN classifier is to find proper features from the given EEG data and feed those feature values to the classifier (i.e., parameterized EEG data as input) . Artificial Neural Network has been used for automated diagnosis by several research groups~\cite{4167902,lvq,A_neural-network-based_detection_of_epilepsy,1028922}.  But those work focus on seizure detection only. Since the interictal EEG does not have seizure activies, the features identified for those neural networks might not work for our purpose: diagnosing epilepy and localizing foci from interictal data. We use four classes of features, namely, power spectral feature, fractal dimensions, Hjorth parameters, and amplitude statistics.

\setlength{\unitlength}{0.4cm}
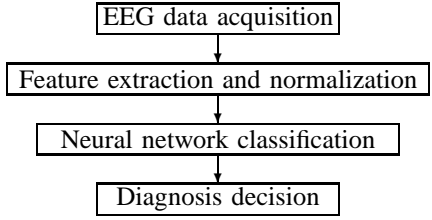
\begin{figure}[!hbt]
\begin{center}
 \begin{picture}(6,6.5)
\put(-1,0){\framebox(8,1){Diagnosis decision}}
\put(3,2){\vector(0,-1){1}}
\put(-3,2){\framebox(12,1){Neural network classification}}
\put(3,4){\vector(0,-1){1}}
\put(-4,4){\framebox(14,1){Feature extraction and normalization}}
\put(3,6){\vector(0,-1){1}}
\put(-1,6){\framebox(8,1){EEG data acquisition}}
\end{picture}
\caption[fig]{Flow diagram of our EEG classification
scheme\label{diagram}}
\end{center}
\end{figure}

Based on a widely used data set with epileptic and normal EEG data, our experiments indicate that interictal EEG of epileptic patients  can be differentiated from those of healthy people with high accuracy and fast speed. Our interictal EEG based diagnostic approach achieves a 99.6\% overall accuracy in cross-validation. The exisiting ictal data based strategy is also tested in our classifier, with 98.3\% accuracty.  Focus localization is achieved with a 78.5\% accuracy. Our classifier is also capable of distinguishing interictal and ictal EEG and thus detecting seizures. The 96.9\% accuracy underlines the possible patient monitoring. The speed of our classifier is very fast -- $0.01$ second per run in all four classification problems. These results imply the possibility of our system for real-time diagnosis, monitoring and focus localization. 

\section{Data Acquisition}
\label{data}
In our experiments, we adopt the data set, which is widely used in previous epileptic diagnosis/analysis research~\cite{A_neural-network-based_detection_of_epilepsy, BOnn_PRE,jnm2005,4167902}, from Klinik f\"{u}r Epileptologie, Universit\"{a}t Bonn, German~\cite{BOnn_PRE}.  It consists of five sets, each containing 100 single-channel EEG segments. Each segment has 4096 sampling points over 23.6 seconds. Note that artifacts, e.g., due to hand or eye movements, have been manually removed by the creators of the data. Data in sets A and B is extracranial EEG from healthy volunteers with eyes open and eyes closed respectively. Sets C and D are intracranial data over interictal period while Set E over ictal period. Segments in D are from within the epileptogenic zone, and those in C from the hippocampal formation of the opposite hemisphere of the brain. All EEG signals were sampled at a sampling rate of 173.61Hz. Refer to \cite{BOnn_PRE} for detailed information of the data. The data was filtered by a low-pass filter of cutoff frequency 40Hz.

\section{Feature Extraction}
\label{feature_extraction}
Our classifier uses 38 features of 4 classes to characterize interictal EEG signal. The power spectral features describe energy distribution in the frequency domain. Fractal dimensions outline the fractal property. Hjorth parameters describe the chaotic behavior. Mean and standard deviation represent the amplitude statistics. Since normalization is very important to distance-based classifier, features are normalized before fed into PNN.

\subsection{Power Spectral Features}
To a time series $x_{1}, x_{2}, \cdots, x_{N}$, its Fast Fourier Transform (FFT) $X_{1}, X_{2}, \cdots, X_{N}$ are estimated as 
$$
X_{k} = \sum_{1}^{N}x_{n}W_{N}^{kn},\ \ k = 1,2,\cdots,N
$$
where $W_{N}^{kn}= e^{\frac{-j2\pi kn}{N}}$ and $N$ is the series length.

\begin{figure}[!hbt]
\begin{center}
\includegraphics[scale=0.26]{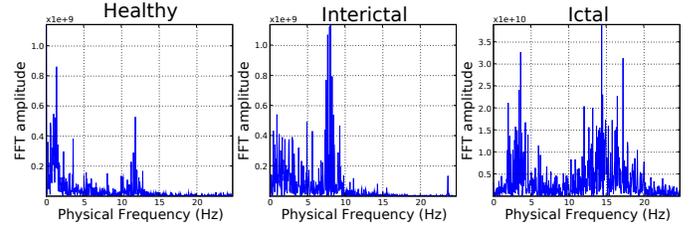}
\caption[fig]{Typical FFT results of 3 EEG segments (Raw data in $\mu$V. For FFT results,  the scale in the Y-axis of ictal data is 10 times larger than the ones of healthy and interictal data)}
\label{FFT}
\end{center}
\end{figure}

We noted a clear difference that in general the ictal EEG has more power components in the higher frequency region ($>$14Hz) while non-ictal EEG are mostly below 14Hz. This point has also been described in \cite{376750}. From Fig. \ref{FFT}, it can be clearly seen that the central frequencies of peaks of different EEG signals lie in different regions.

Based on the FFT result, Power Spectral Intensity (PSI) and Relative Intensity Ratio (RIR) are evaluated to each 2-Hz frequency band from 2-32~Hz. The PSI is defined as 
$$
PSI_{k}=\sum_{i=\lfloor N\frac{f_{min}}{f_{s}}\rfloor}^{\lfloor N\frac{f_{max}}{f_{s}}\rfloor}X_{i}, \ \ \  k=1,2,\dots,15
$$
where $f_{min}=2k$, $f_{max}=2k+2$, $f_s$ is the sampling frequency and $N$ is the series length. As you can see, the $f_{min}$ and $f_{max}$ are the lower and upper boundaries of each 2-Hz band, respectively.
The RIR is defined as 
$$
RIR_{j}=\frac{PSI_{j}}{\sum_{k=1}^{15}PSI_{k}}, \ \ \ j=1,2,\cdots,15 
$$
So we have 15 PSIs and 15 RIPs. They are the first 30 features we used. 

The 2-32Hz band covers some EEG abnormalities unique to epilepsy \cite{waveletEEG}, such as the 3Hz and 6Hz spike waves \cite{Alving}.

\subsection{Petrosian Fractal Dimension (PFD)}
\begin{figure}[!hbt]
\setlength{\abovecaptionskip}{2pt}
 \begin{center}
 \includegraphics[scale=0.43]{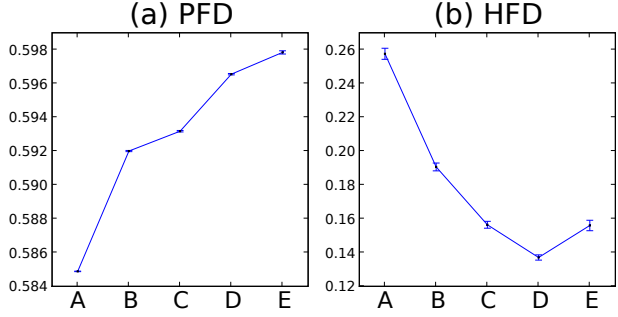}
 \caption[fig]{Average Petrosian Fractal Dimension (a) and Higuchi Fractal Dimension (b) for each set. Error bars denote one standard deviation ($\sigma$) for each class of EEG.}
 \label{PFD}
 \end{center}
 \end{figure}

PFD is defined as:
$$
\text{PFD} = \frac{\log_{10}{N}}{\log_{10}{N} + \log_{10}(\frac{N}{n+0.4N_{\delta}})}
$$
where $N$ is the series length and $N_{\delta}$ is the number of sign changes in the signal derivative~\cite{Petrosian}.
According to Fig.~\ref{PFD}~(a), PFD is highly concentrated within each class and there is no overlap among the data for each class either. Therefore, all classes can be clearly distinguished using PFD.

\subsection{Higuchi Fractal Dimension (HFD)} 
Higuchi's algorithm~\cite{Higuchi} constructs $k$ new series from the original series $ x_{1}, x_{2}, \cdots, x_{N}$ by 
\begin{equation}
x_{m},x_{m+k},x_{m+2k},\cdots, x_{m+\lfloor \frac{N-m}{k} \rfloor k}
\label{Higuchi_series}
\end{equation}
 
where $m = 1, 2, \cdots, k$.

For each time series constructed from (\ref{Higuchi_series}), the length $L(m,k)$ is computed by 
$$
L(m,k) = \frac{\sum_{i=2}^{\lfloor \frac{N-m}{k} \rfloor}|x_{m+ik}-x_{m+(i-1)k}| (N-1)}{\lfloor \frac{N-m}{k} \rfloor k}
$$

The average length $L(k)$ is computed as 
$$
L(k) = \frac{\sum_{i=1}^{k}L(i,k)}{k}
$$

This procedure repeats $k_{max}$ times for each $k$ from 1 to $k_{max}$, and then uses a least-square method to determine the slope of the line that best fits the curve of  $\ln(L(k))$ versus $\ln(1/k)$. The slope is the Higuchi Fractal Dimension. In this paper, $k_{max}$ is 5.

Fig.~\ref{PFD}~(b) indicates HFD is also densely clustered within each class and there is a small overlap between classes C and E. HFD, therefore, is a good feature to characterize classes A, B and E.

\subsection{Hjorth Parameters}
To a time series $x_{1}, x_{2}, \cdots, x_{N}$, the Hjorth mobility and complexity \cite{889990} are respectively defined as  $$\sqrt{\frac{M2}{TP}}$$ and $$\sqrt{\frac{M4\cdot TP} {M2 \cdot M2}}$$ where $TP = \sum x_{i}/N$,  $\ M2 = \sum d_{i}/N$,  $M4 = \sum(d_{i}-d_{i-1})^{2}/N$ and $d_{i}=x_{i}-x_{i-1}$.

\begin{figure}[!hbt]
\begin{center}
\includegraphics[scale=0.46]{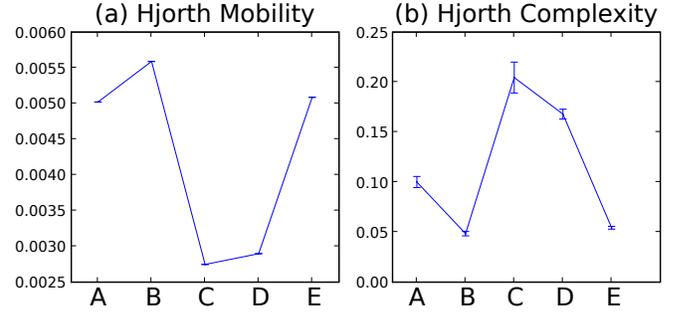}
\caption[fig]{Average Hjorth Mobility (a) and normalized Hjorth Complexity (b) for each set. Error bars denote one standard deviation ($\sigma$) for each class of EEG.}
\label{Hjorth}
\end{center}
\end{figure}

According to Fig. \ref{Hjorth}~(a), Hjorth mobility has a tight distributions within each class. Even though the Hjorth complexity appears very inconsistent among classes, since PNN uses normalized features, we compute the normalized Hjorth complexity and find it also has a tight distribution within each class.

An interesting finding is that a set with low Hjorth mobility would have high normalized Hjorth complexity. For example, Sets C and D have the lowest mobility and highest complexity, which separates them from other sets.

\subsection{Means and standard deviations}
EEG signals from different conditions have different amplitudes. For example, the amplitude of normal activities ranges around 100$\mu$V in this paper while the ictal EEG ranges around 1000$\mu$V. Means and standard deviations of both original data and the absolute values of EEG are evaluated. They are the last 4 features we used.

\section{Probabilistic Neural Network}
\label{PNN}
An AI-based classifier is essentially a mapping $f: \mathbb{R}^{m} \rightarrow \mathbb{Z}^{n}$ from the feature space to the discrete class space. An Artificial Neural Network (ANN) implements such a mapping by using a group of  interconnected artificial neurons simulating human brain. An ANN can be trained to achieve expected classification results against the input and output information stream, so there is not a need to provide a specified classification algorithm. 

PNN  is one kind of distance-based ANNs, using a bell-shape activation function. This technique makes decision boundaries nonlinear and hence it can approach the Bayesian optimal~\cite{specht1988}. Compared with traditional back-propagation (BP) neural network, PNN is considered more suitable to medical application since it uses Bayesian strategy, a process familiar to medical decision makers~\cite{MedPNN}. The real-time property of PNN is also crucial to our research. In PNN, decision boundaries can be modified in real-time as new data become available~\cite{PNN}. There is no need to train the network over the entire data set again. So we can quickly update our network as more and more patients' data becomes available.

Our PNN has three layers: the Input Layer, the Radial Basis Layer which evaluates distances between input vector and rows in weight matrix, and the Competitive Layer which determines the class with maximum probability to be correct. The network structure is illustrated in Fig. \ref{structure}, using symbols and notations in
\cite{NNdesign}. 
 Dimensions of arrays are marked under their names.
 
\setlength{\unitlength}{0.5cm}
 \begin{figure}[!htb]
\begin{center}
 \begin{picture}(20,9.5)
\put(1,0){\oval(1.5,1)[b]}
\put(1,8){\oval(1.5,1)[t]}
 \put(-0.5,9){Input Layer}
 \put(1,1.5){\makebox(0,0){\small{\textit{R}}}}
 \put(1,3){\makebox(0,2){\rule{2mm}{20mm}}}
 \put(1,5){\vector(1,0){2}}
 \put(2,5){\makebox(0,0.6){\textbf{p}}}
 \put(2,5){\makebox(0,-0.7){\small{\small{\textit{1}}$\times$\textit{R}}}}
\put(6,0){\oval(8,1)[b]}
\put(6,8){\oval(8,1)[t]}
\put(4,9){Radial Basis Layer}
 \put(2.3,1){\makebox(0,0){1}}
 \put(2.5,1){\vector(1,0){1}}
 \put(3.5,0.5){\framebox(2,1){\textbf{b}}}
 \put(3.5,7){\framebox(2,1){$ \mathbf{W} $}}
 \put(3.5,6.5){\makebox(0,0){\small{\textit{Q}}$\times$\small{\textit{R}}}}
\put(3,4.5){\framebox(3,1){$\| \mathbf{W} - \mathbf{p}\|$}}
\put(4.5,7){\vector(0,-1){1.5}}
\put(4.5,0){\makebox(0,0){\small{\textit{Q}}$\times$\small{\textit{1}}}}
\put(4.5,3){\circle{1}}
\put(4.4,3){\makebox(0,0){\large{ $\cdot \times$}}}
\put(4.5,4.5){\vector(0,-1){1}}
\put(4.5,1.5){\vector(0,1){1}}
\put(5,3){\vector(1,0){2}}
\put(6,3.5){\makebox(0,0){$\mathbf{n}$}}
\put(6,2.5){\makebox(0,0){\small{\textit{Q}}$\times$1}}
\put(7,1){\framebox(2,4){\epsfig{file=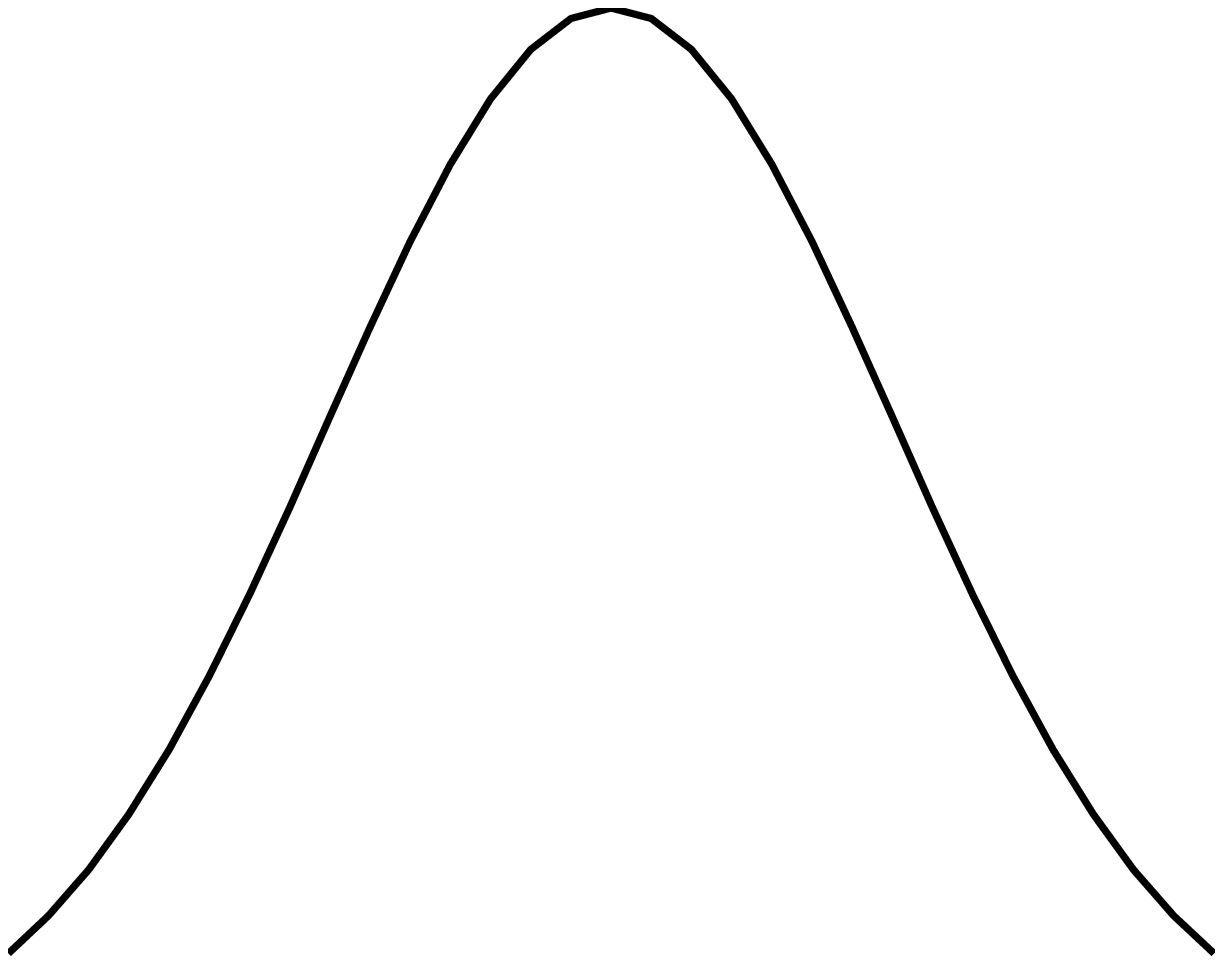,width=29pt,height=30pt}}}
\put(8,1.9){\makebox(0,-0.5){radbas}}
\put(8,0.5){\makebox(0,0){\small{\textit{Q}}}}
\put(9,3){\vector(1,0){2}}
\put(10,3.5){\makebox(0,0){$\mathbf{a}$}}
\put(10,2.5){\makebox(0,0){\small{\textit{Q}}$\times$1}}
\put(11,2.5){\framebox(2,1){$ \mathbf{M} $}}
\put(11.2,1.9){\small{\textit{K}}$\times$\small{\textit{Q}}}
\put(13,3){\vector(1,0){2}}
\put(14,3.5){\makebox(0,0){$\mathbf{d}$}}
\put(14,2.5){\makebox(0,0){\small{\textit{K}}$\times$1}}
\put(15,1){\framebox(1,4){\Large{C}}}
\put(15.5,0.5){\makebox(0,0){\small{\textit{K}}}}
\put(16,3){\vector(1,0){2}}
\put(17,3.5){\makebox(0,0){$\mathbf{c}$}}
\put(17,2.5){\makebox(0,0){\small{\textit{K}}$\times$1}}
\put(14,0){\oval(7,1)[b]}
\put(14,8){\oval(7,1)[t]}
\put(11,9){Competitive Layer}
\end{picture}
\caption{PNN structure, R: number of features, Q: number of training samples, K: number of classes. The input vector $\mathbf{p}$ is presented here as a black vertical bar.}
\label{structure}
\end{center}
\end{figure}


\subsection{Radial Basis Layer}
In Radial Basis Layer, the vector distances between input vector $\mathbf{p}$ and the weight vector made of each row of weight matrix $\mathbf{W}$ are calculated. Here, the vector distance is defined as the dot product between two vectors\cite{specht1988}. The dot product between $\mathbf{p}$ and the $i$-th row of $\mathbf{W}$ produces the $i$-th element of the distance vector matrix, denoted as $||\mathbf{W}  - \mathbf{p} ||$. The bias vector $\mathbf{b}$ is then combined with $||\mathbf{W} - \mathbf{p}||$ by an element-by-element multiplication, represented as ``$\cdot \times$" in Fig. \ref{structure}. The result is denoted as $\mathbf{n} = ||\mathbf{W} - \mathbf{p}|| \cdot \times \mathbf{b}$.

The transfer function in PNN has built into a distance criterion with respect to a center. In this paper, we define it as
\begin{equation}
 radbas(n) = e ^{-{n^2}}
\label{radbas}
\end{equation}
Each element of $\mathbf{n}$ is substituted into (\ref{radbas}) and produces corresponding element of $\mathbf{a}$, the output vector of Radial Basis Layer. We can represent the \textit{i}-th element of $\mathbf{a}$ as 
\begin{equation}
 \mathbf{a}_{i}=radbas(||\mathbf{W}_{i} - \mathbf{p}||\cdot \times \mathbf{b}_{i})
\end{equation}
where $\mathbf{W}_{i}$ is the $i$-th row of $\mathbf{W}$ and $\mathbf{b}_{i}$ is the $i$-th element of bias vector $\mathbf{b}$.

\subsubsection{Radial Basis Layer Weights} Each row of $\mathbf{W}$ is the feature vector of one trainging sample. The number of rows equals to the number of training samples.

\subsubsection{Radial Basis Layer Biases} All biases in radial basis layer are set to $\sqrt{\ln{0.5}}/s$ resulting in radial basis functions that cross 0.5 at weighted inputs of $\pm s$, where $s$ is the spread constant of PNN. In this paper, $s$ is set to 0.1,  since our experiments show the highest accuracy is achieved when $s=0.1$, as illustrated in Fig.~\ref{spread}.

\subsection{Competitive Layer}
There is no bias in Competitive Layer. In this layer, the vector $\mathbf{a}$ is first multiplied by layer weight matrix $\mathbf{M}$, producing an output vector $\mathbf{d}$. The competitive function $\mathbf{C}$ produces a 1 corresponding to the largest element of $\mathbf{d}$, and 0's elsewhere. The index of the 1 is the class of the EEG segment. 
$\mathbf{M}$ is set to $K \times Q$ matrix of $Q$ target class vectors. If the $i$-th sample in training set is of class $j$, then we have a 1 on the $j$-th row of $i$-th column of  $\mathbf{M}$.

\section{Experimental Results}
\label{result}

As shown in Table. \ref{accuracy}, we designed four experiments to test the ability of our classifier to separate:
\begin{enumerate}
 \item normal EEG (sets A and B) and interictal EEG (sets C and D)
 \item normal EEG (sets A and B) and ictal EEG (set E)
 \item interictal EEG (sets C and D) and ictal EEG (set E)
 \item interictal EEG sampled from epileptogenic zone (set C)  and interictal EEG sampled from opposite hemisphere (set D)
\end{enumerate}

The first two experiments evaluate the performance of our algorithm using interictal EEG and ictal EEG respectively. 

The last two experiments evaluate the feasibility of our algorithm on seizure monitoring and focus localization, respectively. 

\begin{table}
\begin{center}
\caption[acc]{Overall accuracy and classification time using PNN\label{accuracy}}
\begin{tabular}[hbt]{  c   c  c  c }
\hline No. & Experiment &  Accuracy & Time (s) \\ 
\hline  & normal (200 samples)  & &  \\
{\raisebox{1ex}[0pt]{1}} & vs. interictal (200 samples) & {\raisebox{1ex}[0pt]{99.5\% }} & {\raisebox{1ex}[0pt]{0.01 }}\\
\hline  & normal (200 samples)  & &   \\
{\raisebox{1ex}[0pt]{2}} & vs. ictal (100 samples) & {\raisebox{1ex}[0pt]{98.3\% }} & {\raisebox{1ex}[0pt]{0.01 }}\\
\hline  & interictal (200 samples)  & &   \\
{\raisebox{1ex}[0pt]{3}} & vs. ictal (100 samples) &  {\raisebox{1ex}[0pt]{96.7\% }} & {\raisebox{1ex}[0pt]{0.01 }}\\
\hline  & epileptogenic zone (100 samples)  &  &  \\
{\raisebox{1ex}[0pt]{4}} & vs. opposite hemisphere (100 samples) &  {\raisebox{1ex}[0pt]{77.5\% }} & {\raisebox{1ex}[0pt]{0.01}} \\
\hline
\end{tabular}
\end{center}
\end{table}

The classifier is validated using leave-one-out cross-validation (LOO-CV) on 400, 300, 300 and 200 samples respectively  in experiments 1, 2, 3 and 4. Our algorithm is implemented using the MATLAB Neural Network Toolbox. Table \ref{accuracy} lists the overall accuracy and classification time of four experiments. 

\begin{figure}[!hbt]
\begin{center}
\includegraphics[scale=0.7]{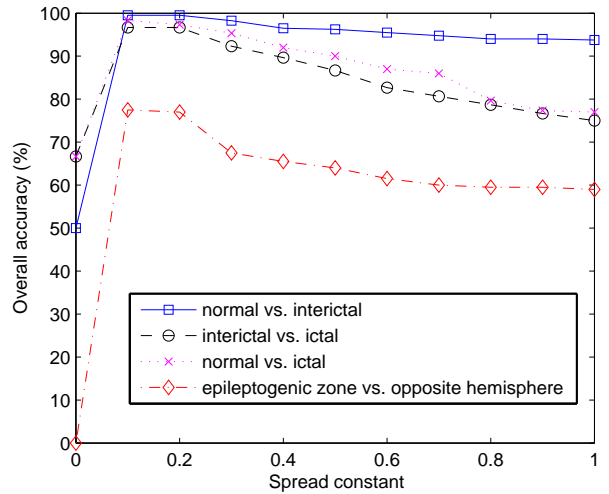}
\caption[fig]{Overall accuracy with respect to PNN spread constant\label{spread}}
\end{center}
\end{figure}

The spread constant of PNN, is seleted according to overall accuracy. As illustrated in Fig.~\ref{spread}, all experiments achieve the highest accuracy, when spread constant is $0.1$. In our experiments, therefore, spread constant is set to $0.1$.

Our approach reaches 99.5\% accuracy in using interictal EEG for epileptic diagnosis (Experiment 1). This result validates the feasibility to use only interictal data in epileptic diagnosis. 

The accuracy of our classifier using ictal data reaches 98.3\%, which is very similar to what has been reported from previous work \cite{4167902}. 

In Experiment 3, 96.7\% accuracy  shows our system can distinguish ictal EEG from interictal EEG very well. This suggests the feasibility to continuously monitor patient status or detect seizures by classifying an EEG segment cut out from monitoring data. If the segment covering current time instant is classified as ictal, then the patient has been in an ictal state, i.e.,  the seizure is occurring. The focus localization experiment achieves a promising accuracy of 77.5\%, which still needs further improvement.

In all the 4 experiments, the classification time per run is 0.01 second (on MATLAB R2008a for Linux, 1.6GHz 64-bit CPU, 2G RAM), which is very short compared with the EEG segment length, 23.6 seconds. This shows the feasibility of real-time monitoring. For long-term monitoring, we can periodically sample the EEG by a sliding window and analyze the windowed segment. For example, immediately after ictal activity detection, devices equipped with our algorithm can send out an alarm to healthcare providers.

\section{Conclusions}
\label{futurework}
In this paper, an automated EEG recognition system for epilepsy diagnosis is developed and validated by cross-validation. Compared with the existing conventional seizure detection algorithms, our approach does not require seizure activity to be captured in EEG recording and thus is “seizure-independent”. This feature relieves the difficulties in EEG collection since interictal data is much easier to be collected than ictal data. 38 EEG features are extracted and PNN is employed to classify those features. 

Experiments indicate that interictal EEG of epileptic patients can be differentiated from those of healthy people with high accuracy and fast speed. Our interictal EEG based diagnostic approach achieves a 99.5\% overall accuracy in cross validation.  Diagnosis based on ictal data is also tested in our classifier, reaching a high 98.3\% accuracy. So, our algorithm works fine with both interictal and ictal data. We also extend the funtion of the classifier, to patient monitoring and focus localization. 96.7\% accuracy is achieved on differentiating ictal from interictal EEG, which suggests the feasibility of online patient monitoring. The focus localization result is also promising with a 77.5\% accuracy. The speed of our classifier is very good, costing only 0.01 second to classify an EEG segment of 23.6 seconds.

\bibliographystyle{IEEEtran}
\bibliography{ictai2008}

\end{document}